\pdfoutput=1

\documentclass[11pt]{article}
\usepackage{multirow}

\usepackage{ACL2023}

\usepackage{times}
\usepackage{latexsym}

\usepackage[T1]{fontenc}

\usepackage[utf8]{inputenc}

\usepackage{microtype}

\usepackage{inconsolata}

\title{Multicultural Name Recognition For Previously Unseen Names}

\author{Alexandra Loessberg-Zahl \\
  Businessolver / Seattle, WA\\
  \texttt{lloessbergzahl@businessolver.com}}

\begin{document}
\maketitle

\begin{abstract}
State of the art Named Entity Recognition (NER) models have achieved an impressive ability to extract common phrases from text that belong to labels such as location, organization, time, and person. However, typical NER systems that rely on having seen a specific entity in their training data in order to label an entity perform poorly on rare or unseen entities \citep{derczynski-etal-2017-results}. This paper attempts to improve recognition of person names, a diverse category that can grow any time someone is born or changes their name.

In order for downstream tasks to not exhibit bias based on cultural background, a model should perform well on names from a variety of backgrounds. In this paper I experiment with the training data and input structure of an English Bi-LSTM name recognition model. I look at names from 103 countries to compare how well the model performs on names from different cultures, specifically in the context of a downstream task where extracted names will be matched to information on file. I find that a model with combined character and word input outperforms word-only models and may improve on accuracy compared to classical NER models that are not geared toward identifying unseen entity values. 
\end{abstract}

\section{Introduction and Background}

Named entity recognition (NER) is a natural language processing (NLP) task in which a computer finds and labels elements in a piece of text that fall into a set of predetermined categories, such as person names, locations, medical terminology, and organization names. This task is an integral part of information extraction from unstructured text.

This paper focusses on just one of those categories - person names. Specifically, I experiment with model input and training data composition to see if I can develop a model that increases true positives when extracting names from text. Although this work could benefit a variety of downstream tasks, it was developed specifically for use with Businessolver’s virtual benefits agent, Sofia, in order to match extracted names with user information on file, such as names of dependents and beneficiaries. If these names are extracted successfully from a user’s utterance, we can then use them to complete requests from the user, such as checking a dependent’s benefits, or viewing beneficiary designations. 

A shared task by \citealp{derczynski-etal-2017-results} presented a curated test dataset of rare and emerging entities to evaluate NER model performance on changing language and novel entities introduced into language over time. The top performing model had an F1 score of 41.86 and the task found that models had little difficulty with common English names but a harder time with identifying names or locations from other languages. In general, the models participating in this task showed that identifying and labeling these rare or novel entities was more difficult than identifying high frequency entities, which typically make up a larger portion of test data in entity recognition tasks that are not geared toward emerging language.

Based on these findings, and because the majority of Sofia’s user base resides in the United States, a culturally diverse country, I aim to create a model that performs equally well across names from various cultural backgrounds and does not exhibit performance bias between common and rare names. I believe that using a model that incorporates character-level input could be beneficial, especially in reducing the effects of out of vocabulary names not present in the training data.

In terms of model architecture, there are many existing approaches to NER. Yadav and Bethard \citeyearpar{yadav-bethard-2018-survey} summarize the highest performing models in 4 categories; knowledge-based systems (which use lexicons and domain specific resources and therefore do not need annotated training data), unsupervised and bootstrapped systems (which use cues like capitalization, extracted patterns, locations of noun phrases, etc.), feature-engineered supervised systems (which use annotated training data and frameworks such as HMMs, CRFs, SVMs, and decision trees as well as features such as orthography, presence of certain prefixes or suffixes, location in the sentence, trigger words, etc.), and feature-inferring neural network systems (which give either words, characters, or both as input to a recurrent neural network (RNN), and use pre-trained word and character embeddings.

Of these approaches, the best F score for English was acheived by an RNN with character and word level architecture from Chiu and Nichols \citeyearpar{chiu-nichols-2016-named}, while the best F scores for Spanish, Dutch, and German were obtained by an RNN with a character, word, and affix architecture by Yadav et al. \citeyearpar{yadav-etal-2018-deep}. These results were obtained using datasets from the CoNLL02 and CoNLL03 shared tasks \citep{tjong-kim-sang-2002-introduction} \citep{tjong-kim-sang-de-meulder-2003-introduction}. Another shared task, DrugNER, evaluated NER results on a corpus of medical and drug terminology. On this task, the word and character level architecture from Yadav et al. \citeyearpar{yadav-etal-2018-deep} performed better than just word level architecture alone from Chalapathy et al. \citeyearpar{chalapathy-etal-2016-investigation}.

This paper will compare models trained on combined word and character input and those trained solely on word input. Based on previous studies and the intuition that character level input can allow the model to learn sub-word character patterns that can be used to identify commonalities between names and contexts seen in training and those seen the test data or real world applications, I hypothesize that the word and character level architecture will produce better performing models. For example, a model with character level input might have a better chance of identifying names that are similar to names seen in the training data, such as “Ashleigh” and “Ashley”, “Alex” and “Alexis”, or “Drew” and “Andrew”, since common character patterns are present between these pairs of names.

The data curated for this project aims to minimize bias toward common names. This is achieved by allowing each name in the training set to appear only once. By removing familiarity as a cue for the model to train on, the goal is to create a model that relies more on sentence context to recognize names, rather than how often the name was seen in training. I also source names from a variety of cultures to expose the model to different character patterns and to evaluate performance across countries of origin. 

Throughout this paper, I will use the word “names” which can be confused with names for other entity instances, or the named entity recognition task itself. Unless otherwise specified, the word “names” will only be used to refer to person names. 

In the sections that follow I will discuss the composition and sources of the training and test data for this project, the model architectures used, and the results and conclusion.

\section{Methods}

\subsection{Data}

The training and test data for the model consist of unique utterances which all contain unique person names. That is, if you were to remove the names from each utterance, no two utterances would be the same, and very few utterances would contain the same name or have any overlap in first or last name. This holds true both within and across the train and test sets. The goal of this data composition is to train a model that learns generalized patterns across names and utterances and does not rely on memorization of either individual names or exact utterances in order to extract names. The sections below will describe the sources and composition of the data used in this project.

\subsubsection{Utterances}

The training set for the model consists of 59,663 unique utterances that come from two main sources. 41,376 utterances come from Businessolver’s proprietary training set for our virtual agent, Sofia. Because Sofia is a virtual agent in the employee benefits domain, the majority of the utterances in this training set are benefits related (i.e. relating to medical, dental, vision, and other types of insurance, as well as employee programs and resources). These utterances were sourced first from the subset of Sofia’s training set tagged with a person name entity already. However, there were only 194 utterances with both unique names AND unique utterances in that subset of Sofia’s current training set. In order to pull more benefits related utterances, I also sourced utterances from Sofia’s training set that contain one of our dependent-type entities and proceeded to replace those entities with names. These entities consist of family members who might be covered on a user’s insurance, such as “spouse”, “wife”, “husband”, “domestic partner”, “child”, “daughter”, and “son”. Utterances containing these entities are good candidates for augmenting the model’s training data because, like names, they also refer to a person. Replacing one of these entities with a name results in a natural sounding utterance. For example, in the utterance “My daughter just turned 26, and is no longer eligible to be on my insurance” we would replace “my daughter” with a name like “Erica Gupta” resulting in the new utterance “Erica Gupta just turned 26, and is no longer eligible to be on my insurance.” 41,182 of these dependent-tagged utterances were sampled for the training data. The source of the names and process followed to insert them in these utterances are described in section 2.1.2 below. 

To add some variety in utterance topics, 18,287 additional training set utterances containing person names were sourced from an entity annotated corpus on Kaggle \citep{NERDataset2021} which contains sentences from the Groningen Meaning Bank \citep{bos-2013-groningen} of public domain English text (newspaper articles, etc.). In cases where any of these sentences contained names already pulled from Sofia’s training data, the names were removed from the sentences and replaced with unique names, as described in section 2.1.2.

The utterances in the test set were pulled from all the same sources as the training set, using the same methods of replacing names as necessary (see section 2.1.2). It contains 5,029 total utterances, 3,086 of which were sourced from Businessolver’s proprietary NLU training data, consisting of examples tagged with person entities as well as dependent type entities. 1,943 more utterances were pulled from the Groningen Meaning Bank corpus, again pulling only examples tagged with person entities.

As with the training set, the test set contains only unique utterances and names, with no utterances or names overlapping with the training set. That is, no utterances or names in the test set were seen in the training set, which allows us to test the model’s performance on unique unseen examples and names, simulating real world scenarios in which a user’s utterance contains a name or utterance that the model hasn’t been trained on.

\subsubsection{Names}

The unique names used to replace those in the Groningen corpus and the dependent entities from Sofia’s training data come from two sources. First, for the training data, 44,989 names come from a random sample of Businessolver’s database of user information. Because these are the names of actual users, they are a good representation of the types of names this model will be used to recognize in its downstream implementation. In the test set, only 95 names were from the Businessolver user information database in order to down sample the number of names from the US compared to other countries.

While the names from Businessolver’s user base are representative of the names this model will come across in production, the vast majority of our users reside in the United States. Because we want to test if this model can perform well on names from various cultural backgrounds, I also sourced names from the name-dataset python package \citep{NameDataset2021} which contains names from a Facebook data dump and provides first and last names from 105 different countries, sorting based on how frequent each name is in the set of names from that country. This package also allows you to look at a certain gender when selecting names, although it only supports binary gender labels, with no label option for non-binary or gender non-conforming. 

In order to select names from this dataset, 200 female first names, 200 male first names, and 200 last names were pulled for each country in the dataset. Half of the first names were randomly assigned a last name, resulting in 100 female names with last names, 100 male names with last names, and 100 of each female and male first names with no last names. Those names were then reduced, eliminating names containing special characters and diacritics that are not supported by the model. Because this data comes from different countries, many names in the dataset are written in non-Latin orthography, and these were excluded as well due to the domain of the task (i.e. an English based virtual assistant).  Furthermore, because no duplicate first or last names were permitted in the training set, the pool of names was reduced even further due to duplicate names across different countries. The resulting quantities of names across countries were not as balanced as I would have liked, even despite pulling additional names past the top 200 from each country. However, we can still compare results across countries, and future research may explore other name sources that allow for more even numbers across countries. For example, datasets of transliterated names from non-Latin languages would be valuable in increasing coverage of names from cultures that predominantly use non-Latin script.

The resulting set of names were used to fill utterances in both the training and test set. 13,936 of these names were used in the training set, and 4,929 of these names were used in the test set. See Appendix A for the breakdown of names in the training set and test set based on the country they were sourced from. 

Of the names in the training set 37,312 were female names, 21,459 were male names, and 892 did not have an associated gender. Those names without associated gender came primarily from Sofia’s training set and names from Businessolver’s user base that did not have a gender labeled. In the test set, 3,058 names were female, 1,968 were male, and 5 did not have an associated gender.

It is worth noting that the final training and test sets did not include any utterances that do not contain a name. In an early experiment, a model was trained on examples with and without names, and recall (true positives) dropped significantly. Due to this result, and because this data was developed for a downstream task where maximizing true positives is more valuable than minimizing false positives, we decided to move forward with training and test sets that do not contain any examples without names.

In both the training and test sets, each utterance has only one name tagged. However, some of the sourced utterances have the possibility of containing multiple names. During preprocessing, any names that exist elsewhere in the training set are tagged when they occur in utterances with multiple names. For example, if we have an utterance like “Sarah and Mike both need glasses” and the name tagged for that utterance is “Mike” – if “Sarah” is tagged in a different utterance in the data, we add an additional tag to this utterance so both names are tagged. In this way, we do get a minimal number of duplicate names tagged in the training data, but still no overlap in names tagged in the train and test sets. 

\subsection{Model Architecture}

Two main model architectures are compared in this research: A Bidirectional LSTM that takes only word-level input, and a Bidirectional LSTM that takes both word-level and character-level input. In the word-only model, each word of each utterance is mapped to a corresponding integer and fed into a Keras embedding layer resulting in a dense vector. In the word and character model, the words are embedded the same way, and the characters of each word are mapped to a corresponding integer and encoded through a time distributed LSTM with 20 hidden memory units.

In the word and character model, after the encoding, the resulting word and character embeddings are concatenated such that each word embedding is concatenated with the character embedding representing the character sequence that makes up that word.

Beyond the input encoding step, the two models are identical. The hidden bidirectional layers, one forward, one backward, have 50 memory units each and the output from these layers is concatenated. The output layer uses a softmax activation function and is wrapped in a time distributed layer in order to predict one value per word in the utterance. Both models use ADAM optimization and sparse categorical cross entropy as the loss function.

To account for the stochastic nature of these models, 5 models were trained for both the character + word architecture and the word only architecture, and evaluated on the test set. This allows us to compare results between the two model architectures across multiple unique models to see if one architecture consistently outperforms the other. 

\section{Evaluation}

Each model is evaluated on the test set, measuring both strict accuracy and partial accuracy. In the strict accuracy measure, true positives are calculated as the number of names extracted completely correctly, that is, if the name has a first and last name, both parts of the name must be extracted correctly. Partial accuracy allows names where only one part of the name (either the first name or last name) is extracted correctly. In order to count toward partial accuracy, the extracted name cannot contain any words that are not part of the full name. Take for example the sentence “I need to add medical coverage for my daughter Kelsey Scott so she can see the doctor”. If the model extracts the name “Kelsey Scott”, we count this as a fully correct true positive. If the model extracts only “Kelsey” or only “Scott”, we count this as a partially correct answer. But if the model extracts “daughter Kelsey” or “Kelsey Scott so”, these are not counted as true positives for either accuracy measure because they contain words that are not part of the name. 

I do not measure precision, recall, and F1 because this paper focuses on how well the models extract names when there is a name in the utterance (true positives). Furthermore, these standard NER metrics don’t make as much sense for these models as they do for other NER models since the training and test sets do not include any utterances without person names. As mentioned in section 2.1.2, an early experimental model trained on examples with and without names saw a significant decrease in the proportion of true positives. For the specific downstream task this model was developed for, where extracted names will be validated against user information on file, maximizing true positives is more valuable than minimizing false positives. Therefore, we rely on the strict and partial accuracy measures described above, and acknowledge that the resulting models will have a high likelihood of false positives.

\section{Results and Discussion}

The strict accuracy and partial accuracy for each of the 5 models trained for either architecture – word and character, or word only – are shown in Table \ref{tab:accuracy-scores} along with the average of these scores for each architecture. 

\begin{table*}[]
\begin{tabular}{cllcll}
\hline
\multicolumn{3}{c|}{\textbf{Word + Character Model}}                                            & \multicolumn{3}{c}{\textbf{Word Only Model}}                                                    \\ \hline
\multicolumn{1}{l}{\textbf{Model Number}} & \textbf{Accuracy Measure} & \textbf{Score} & \multicolumn{1}{l}{\textbf{Model Number}} & \textbf{Accuracy Measure} & \textbf{Score} \\
\multirow{2}{*}{1}                        & Strict Accuracy           & 0.8008         & \multirow{2}{*}{1}                        & Strict Accuracy           & 0.7451         \\
                                          & Partial Accuracy          & 0.8622         &                                           & Partial Accuracy          & 0.8566         \\
\multirow{2}{*}{2}                        & Strict Accuracy           & 0.8087         & \multirow{2}{*}{2}                        & Strict Accuracy           & 0.7411         \\
                                          & Partial Accuracy          & 0.8618         &                                           & Partial Accuracy          & 0.8646         \\
\multirow{2}{*}{3}                        & Strict Accuracy           & 0.8117         & \multirow{2}{*}{3}                        & Strict Accuracy           & 0.7290         \\
                                          & Partial Accuracy          & 0.8662         &                                           & Partial Accuracy          & 0.8550         \\
\multirow{2}{*}{4}                        & Strict Accuracy           & 0.8004         & \multirow{2}{*}{4}                        & Strict Accuracy           & 0.7300         \\
                                          & Partial Accuracy          & 0.8570         &                                           & Partial Accuracy          & 0.8465         \\
\multirow{2}{*}{5}                        & Strict Accuracy           & 0.8065         & \multirow{2}{*}{5}                        & Strict Accuracy           & 0.7258         \\
                                          & Partial Accuracy          & 0.8620         &                                           & Partial Accuracy          & 0.8475         \\
\multirow{2}{*}{Average}                  & Strict Accuracy           & \textbf{0.8056}         & \multirow{2}{*}{Average}                  & Strict Accuracy           & \textbf{0.7242}         \\
                                          & Partial Accuracy          & \textbf{0.8618}         &                                           & Partial Accuracy          & \textbf{0.8540}        
\end{tabular}
\caption{Strict and partial accuracy scores across each of the 5 model trains for the two model types (word and character, or word only). The last row gives the average scores across all model trains for each of the model types.}
\label{tab:accuracy-scores}
\end{table*}

\begin{table*}[]
\begin{tabular}{lcccc}
                                                  & \multicolumn{2}{c}{\textbf{Only First Names}}                                          & \multicolumn{2}{c}{\textbf{First and Last}}                                            \\ \cline{2-5} 
                                                  & \multicolumn{1}{l}{\textbf{Raw Count}} & \multicolumn{1}{l}{\textbf{Percent Accuracy}} & \multicolumn{1}{l}{\textbf{Raw Count}} & \multicolumn{1}{l}{\textbf{Percent Accuracy}} \\ \cline{1-5} 
\multicolumn{1}{l|}{\textbf{First Name Accuracy}} & 1741/1994                              & 87.22\%                                        & 2702/3033                              & 89.09\%                                        \\
\multicolumn{1}{l|}{\textbf{Last Name Accuracy}}  & --                                     & --                                            & 2588/3033                              & 85.32\%                                       
\end{tabular}
\caption{Accuracy scores for the top performing word + character model on first and last names. The “Only First Names” column shows performance on utterances that contain names with no last name. The “First and Last” column shows performance on utterances that contain names with both a first and last name.}
\label{tab:word-char-first-last}
\end{table*}

\begin{table*}[]
\begin{tabular}{lcccc}
                                                  & \multicolumn{2}{c}{\textbf{Only First Names}}                                          & \multicolumn{2}{c}{\textbf{First and Last}}                                            \\ \cline{2-5} 
                                                  & \multicolumn{1}{l}{\textbf{Raw Count}} & \multicolumn{1}{l}{\textbf{Percent Accuracy}} & \multicolumn{1}{l}{\textbf{Raw Count}} & \multicolumn{1}{l}{\textbf{Percent Accuracy}} \\ \cline{1-5} 
\multicolumn{1}{l|}{\textbf{First Name Accuracy}} & 1727/1994                              & 86.52\%                                        & 2697/3033                              & 88.92\%                                        \\
\multicolumn{1}{l|}{\textbf{Last Name Accuracy}}  & --                                     & --                                            & 2545/3033                              & 83.91\%                                       
\end{tabular}
\caption{Accuracy scores for the top performing word only model on first and last names. The “Only First Names” column shows performance on utterances that contain names with no last name. The “First and Last” column shows performance on utterances that contain names with both a first and last name. }
\label{tab:word-first-last}
\end{table*}

\begin{table}[]
\begin{tabular}{ll}
\hline
\textbf{Measure}        & \textbf{Result} \\ \hline
Overall Strict Accuracy & 0.8117          \\
F Strict Accuracy       & 0.8260          \\
M Strict Accuracy       & 0.7889         
\end{tabular}
\caption{Accuracy scores for the top performing word + character model on both female and male names.}
\label{tab:gender}
\end{table}

Across all model trainings, the word + character models outperformed all of the word only models on the strict accuracy score. When strict accuracy is averaged across all five models in each architecture, the word + character model average accuracy is 7.14\% higher than the average strict accuracy score for the word only model. On partial accuracy, results are more mixed, but on average word + character models outperformed the word only model partial accuracy by 0.77\%. These results support the hypothesis that a model that uses both character and word level input is better able to identify names in sentences. This is likely due to the additional context provided by character embeddings, which can allow a model to apply knowledge about words it has seen in training to novel contexts that may have some similarity in word forms, even if not all of those exact words were in the training data. For example, if we have “Alexis sent in the form” In the training data and then “Alexa sends in the form” in the test data – the models trained without character embeddings will treat “sent” and “sends” as two entirely different words. The models trained \textit{with} character embeddings will be able to identify that the two words share three of the same letters, and potentially use that cue to know that the prior word is likely a name. The character models may also be able to identify character patterns within names themselves if there are meaningful patterns that indicate that a word is a pattern. For example, the model may infer that “Alexa” is similar to the name “Alexis” that it saw in training, and therefore tag the former as a name. Of course, we can’t be sure that this is what the model is actually learning, due to the black-box nature of neural nets.

The relatively smaller gain on partial accuracy for the word + character models could be due to the presence of more utterances with only first names in the training data compared to utterances with full names. To explore this hypothesis, tables \ref{tab:word-char-first-last} and \ref{tab:word-first-last} compare the results from the top performing word + character model and the top performing word only model on first names and last names respectively. We see that both models perform quite well on recognizing first names, and both models see a dip in performance on last names compared to first names. This dip is more pronounced for the word only model. In the word only model, the accuracy for first names in utterances containing a full name is around 5\% higher than the accuracy for extracting last names. Comparatively, this difference is 4\% in the word + character model. We can hypothesize that while extracting last names is more difficult than extracting first names for both models (likely due to over representation of first names in the training data) the extra information encoded in character embeddings allows the word + character model to close this gap between first and last names more than the word model.

A training set more balanced in terms of first and last names could potentially neutralize this difference. However, the word + character model still has acceptably high scores for both first and last names to allow us to be confident that the model will perform well regardless of whether a user includes a full name or just a first name in their utterance. 

Returning briefly to the results in table \ref{tab:accuracy-scores}, the decent performance by both word + character models and word only models on the test set – which contains only names that were not seen in the training data – supports the hypothesis that eliminating duplicate names and contexts from the training data allows the model to generalize to unseen names and contexts. 

The following discussion will focus on results from only the top performing word + character model (word + character model 3 in Table \ref{tab:accuracy-scores}). Table \ref{tab:gender} shows that the model has slightly lower accuracy on male names than female names. This may be due to the higher proportion of female name examples in the training set compared to male names. Future work could train on a data set that is more gender balanced to see if it results in neutralized gender performance. 

Table 7 in Appendix A shows both the strict and partial accuracy across all countries for the top performing character + word model. The majority of the countries represented in the test data have strict accuracy scores above 80\% with even higher partial accuracy scores. This high performance shows the promise of this approach in performing well on names from a diverse set of cultural backgrounds and therefore makes it a valuable tool in reducing bias in downstream applications.

Among false positive results – that is, instances where the model extracted words that were not part of names – there were a few patterns to note. Country and city names like Iran, Syria, Russia, Paris, Egypt, and London show up among the most frequent false positives. This could be due to similarities in character patterns between names from different countries and the names of those countries themselves, or sentence contexts, since both names and country names are proper nouns. Another possibility is that the model has learned to recognize the difference between patterns of characters in typical English words and patterns of characters in names. If this is the case it would also explain why the model picks up location names as well as some benefits specific abbreviations like HSA and LUFSA that differ from the orthographic patterns of most English words.

I did not notice any patterns in the results for countries with lower performance, other than that those countries with relatively less representation in the training data had comparatively lower partial and strict accuracy. As mentioned in section 2.1.2, the training set does not have an equal number of examples across all countries of origin, due to the need to eliminate any duplicate names combined with the presence of special characters or non-Latin orthography. Future research could explore name sources that provide transliterated versions of names from countries and languages that predominantly use non-Latin script to allow for more even numbers across countries. 

Overall, these results are a successful experiment in recognizing names for the given use case – extracting novel or unseen names from user utterances in order to return or change information on file for people related to the user (beneficiaries, dependents, etc). In this case, false negatives are not detrimental to downstream performance, especially when the model shows high performance on true positives. However, for other use cases, it could be beneficial to try this same approach with a wide range of entities, not just person names, and see if person names still see the same benefit and whether more label classes will diminish the effect of this benefit for names or any other entity type.

\section{Conclusion}

In this paper I present the result of a comparison between two BiLSTM models on the task of multi-cultural name recognition in short text. Both models are trained on curated data in which neither names nor utterances are seen more than once in the training and there is no overlap in names or utterances between the train and test data. I found that the model that takes both character and word embeddings as input performs better on both partial accuracy and strict accuracy measures, with a larger benefit observed on the strict accuracy score. I conclude that both the addition of character embeddings and the curation of a dataset that does not allow a model to make predictions based on frequency are successful in increasing true positives in the task of recognizing novel and unseen names. Furthermore, the word and character model produced high accuracy across the majority of countries from which names were sourced and evaluated on, indicating that this method works not only for English names, but for names from diverse cultural backgrounds, making it a valuable tool in reducing bias in name recognition applications.

\bibliography{anthology,custom}
\bibliographystyle{acl_natbib}

\appendix

\section{Appendix}
\label{sec:appendix}
See below for additional tables.
\begin{table*}
\centering
\begin{tabular}{lllllc}
\hline
\textbf{Country} & \textbf{Count} & \textbf{Country} & \textbf{Count} & \textbf{Country} & \textbf{Count}\\
\hline
USA & 44,504 & Bangladesh & 140 & Palestine & 47 \\
Unknown & 1,290 & India & 136 & Sudan & 43 \\
Canada & 395 & Cameroon & 136 & Syrian Arab Republic & 43\\ 
Botswana & 262 & Ireland & 136 & Oman & 41\\ 
Azerbaijan & 256 & Sweden & 132 & Peru & 41\\
Japan & 245 & China & 128 & Iraq & 35 \\ 
Fiji & 245 & Bolivia & 126 & Saudi Arabia & 30 \\
Lithuania & 244 & Honduras & 126 \\
Albania & 244 & Guatemala & 117 \\
Burkina Faso & 235 & Tunisia & 116 \\
Austria & 231 & Cambodia & 115  \\
Malta & 230 & Italy & 111 \\
Maldives & 229 & Iran & 109 \\
Czechia & 314 & Costa Rica & 115\\
Estonia & 214 & Netherlands & 108\\
Cyprus & 213 & Bahrain & 106\\
Haiti & 211 & El Salvador & 106 \\
Croatia & 211 & Switzerland & 104 \\
Ecuador & 210 & Panama & 103  \\
Iceland & 210 & Nigeria & 103\\
Denmark & 207 & Algeria & 101 \\
Brunei Darussalam & 206 & Mauritius & 101\\
Ethiopia & 204 & Greece & 100\\
Finland & 203 & Lebanon & 98\\
Angola & 200 & Germany & 98\\
Burundi & 199 & Brazil & 96\\
Poland & 195 & Morocco & 91\\
Georgia & 194 & Israel & 90\\
Indonesia & 192 & Kazakhstan & 83 \\
Turkey & 186 & Russian Federation & 81\\
Hungary & 178 & Hong Kong & 80 \\
Puerto Rico & 171 & United Kingdom & 80\\
Slovenia & 169 & Qatar & 74\\
Luxembourg & 166 & Chile & 67 \\
Turkmenistan & 166 & France & 67 \\
United Arab Emirates & 165 & Macao & 65\\
Argentina & 163 & Uruguay & 64\\
Ghana & 160 & Singapore & 63\\
Djibouti & 156 & Malaysia & 63\\
Serbia & 156 & Colombia & 63 \\
Norway & 153 & Spain & 61\\
Bulgaria & 151 & Mexico & 57\\
Jamaica  & 150 & Yemen & 57\\
Philippines & 146 & Taiwan & 55\\
Moldova & 144 & Jordan & 54\\
South Africa & 144 & Kuwait & 52\\
Belgium & 143 & Egypt & 52\\
Afghanistan & 141 & Libya & 48\\

\hline
\end{tabular}

\caption{The count of names from each country present in the training data.}
\label{tab:training-names}
\end{table*}

\begin{table*}
\centering
\begin{tabular}{lllllc}
\hline
\textbf{Country} & \textbf{Count} & \textbf{Country} & \textbf{Count} & \textbf{Country} & \textbf{Count}\\
\hline
USA & 121 & Jamaica & 50 & Syrian Arab Republic & 26 \\
Botswana & 81 & Ireland & 49 & Oman & 25 \\
Azerbaijan & 78 & Bulgaria & 49 & Iraq & 25\\ 
Albania & 75 & Norway & 49 & Saudi Arabia & 25\\ 
Lithuania & 75 & Afghanistan & 46 & Libya & 25\\
Burkina Faso & 74 & China & 46 & Sudan & 24 \\ 
Fiji & 73 & Moldova & 46 & Yemen & 24 \\
Austria & 73 & El Salvador & 45 \\
Maldives & 71 & Sweden & 44 \\
Japan & 71 & Bahrain & 44 \\
Angola & 69 & Italy & 43  \\
Czechia & 68 & Panama & 43 \\
Malta & 68 & Tunisia & 43 \\
Brunei Darussalam & 68 & Algeria & 42\\
United Arab Emirates & 66 & Brazil & 42\\
Croatia & 66 & Costa Rica & 42\\
Cyprus & 66 & Netherlands & 41 \\
Argentina & 65 & Mauritius & 41 \\
Ethiopia & 65 & Cambodia & 41  \\
Finland & 65 & Nigeria & 40\\
Ecuador & 65 & Switzerland & 40 \\
Burundi & 64 & Morocco & 39\\
Denmark & 64 & Israel & 38\\
Estonia & 64 & Iran & 38\\
Haiti & 63 & Germany & 37\\
Indonesia & 62 & Qatar & 36\\
Iceland & 60 & Lebanon & 36\\
Poland & 60 & Chile & 36\\
Hungary & 59 & United Kingdom & 35 \\
Georgia & 59 & Hong Kong & 34\\
Turkey & 58 & Greece & 34 \\
Canada & 57 & Malaysia & 34\\
Puerto Rico & 57 & Spain & 34\\
Belgium & 56 & Egypt & 34 \\
Luxembourg & 56 & France & 33 \\
Djibouti & 55 & Colombia & 33\\
Bangladesh & 53 & Mexico & 33\\
Slovenia & 53 & Uruguay & 32\\
Ghana & 53 & Peru & 32\\
Philippines & 52 & Singapore & 31 \\
Turkmenistan & 52 & Portugal & 30\\
South Africa & 51 & Kuwait & 30\\
Guatemala  & 51 & Russian Federation & 30\\
Serbia & 51 & Kazakhstan & 29\\
Cameroon & 51 & Jordan & 29\\
Bolivia & 50 & Macao & 28\\
Honduras & 50 & Taiwan & 28\\
India & 50 & Palestine & 26\\

\hline
\end{tabular}

\caption{The count of names from each country present in the test data.}
\label{tab:test-names}
\end{table*}

\begin{table*}[]
\begin{tabular}{llllllll}
\hline
\textbf{Country}  & \textbf{\begin{tabular}[c]{@{}l@{}}Strict \\ Accuracy\end{tabular}} & \textbf{\begin{tabular}[c]{@{}l@{}}Partial \\ Accuracy\end{tabular}} & \textbf{Support} & \textbf{Country}     & \textbf{\begin{tabular}[c]{@{}l@{}}Strict\\ Accuracy\end{tabular}} & \textbf{\begin{tabular}[c]{@{}l@{}}Partial\\ Accuracy\end{tabular}} & \textbf{Support} \\ \hline
Croatia           & 0.9545                                                              & 0.9545                                                               & 66               & Brazil               & 0.8095                                                             & 0.8810                                                              & 42               \\
Greece            & 0.9412                                                              & 0.9412                                                               & 34               & Cambodia             & 0.8049                                                             & 0.9268                                                              & 41               \\
Moldova           & 0.9348                                                              & 0.9348                                                               & 46               & Netherlands          & 0.8049                                                             & 0.8537                                                              & 41               \\
Turkmenistan      & 0.9231                                                              & 0.9231                                                               & 52               & Cameroon             & 0.8039                                                             & 0.9020                                                              & 51               \\
Hungary           & 0.9153                                                              & 0.9492                                                               & 59               & Denmark              & 0.7969                                                             & 0.8906                                                              & 64               \\
Afghanistan       & 0.9130                                                              & 0.9348                                                               & 46               & Georgia              & 0.7966                                                             & 0.8136                                                              & 59               \\
Brunei Darussalam & 0.9118                                                              & 0.9559                                                               & 68               & Bulgaria             & 0.7959                                                             & 0.7959                                                              & 49               \\
Iceland           & 0.9000                                                              & 0.9000                                                               & 60               & Egypt                & 0.7941                                                             & 0.8824                                                              & 34               \\
Switzerland       & 0.9000                                                              & 0.9000                                                               & 40               & United States        & 0.7934                                                             & 0.8512                                                              & 29               \\
Czechia           & 0.8971                                                              & 0.8971                                                               & 68               & Kazakhstan           & 0.7931                                                             & 0.8276                                                              & 28               \\
Malta             & 0.8971                                                              & 0.9412                                                               & 68               & Taiwan               & 0.7857                                                             & 0.8571                                                              & 60               \\
Albania           & 0.8933                                                              & 0.8933                                                               & 75               & Poland               & 0.7833                                                             & 0.8333                                                              & 46               \\
Macao             & 0.8929                                                              & 0.9286                                                               & 28               & China                & 0.7826                                                             & 0.8696                                                              & 57               \\
Ethiopia          & 0.8923                                                              & 0.9077                                                               & 65               & Puerto Rico          & 0.7719                                                             & 0.8947                                                              & 30               \\
Qatar             & 0.8889                                                              & 0.8889                                                               & 36               & Portugal             & 0.7667                                                             & 0.8000                                                              & 34               \\
Philippines       & 0.8846                                                              & 0.9231                                                               & 52               & Hong Kong            & 0.7647                                                             & 0.8824                                                              & 55               \\
South Africa      & 0.8824                                                              & 0.9020                                                               & 51               & Djibouti             & 0.7636                                                             & 0.8545                                                              & 63               \\
Algeria           & 0.8810                                                              & 0.9048                                                               & 42               & Costa Rica           & 0.7619                                                             & 0.8571                                                              & 42               \\
Jamaica           & 0.8800                                                              & 0.9200                                                               & 50               & Haiti                & 0.7619                                                             & 0.8571                                                              & 50               \\
Turkey            & 0.8793                                                              & 0.8966                                                               & 58               & Bolivia              & 0.7600                                                             & 0.8600                                                              & 29               \\
Burkina Faso      & 0.8784                                                              & 0.8919                                                               & 74               & Jordan               & 0.7586                                                             & 0.8276                                                              & 53               \\
Lithuania         & 0.8784                                                              & 0.8784                                                               & 74               & Slovenia             & 0.7547                                                             & 0.7736                                                              & 65               \\
Finland           & 0.8769                                                              & 0.8769                                                               & 65               & Argentina            & 0.7538                                                             & 0.8154                                                              & 56               \\
Fiji              & 0.8767                                                              & 0.8904                                                               & 73               & Luxembourg           & 0.7500                                                             & 0.8393                                                              & 51               \\
Japan             & 0.8732                                                              & 0.8873                                                               & 71               & Guatemala            & 0.7451                                                             & 0.7843                                                              & 66               \\
Azerbaijan        & 0.8718                                                              & 0.8718                                                               & 78               & UAE & 0.7424                                                             & 0.8939                                                              & 30               \\
Morocco           & 0.8718                                                              & 0.8718                                                               & 39               & Russia   & 0.7333                                                             & 0.7667                                                              & 41               \\
Indonesia         & 0.8710                                                              & 0.9032                                                               & 62               & Mauritius            & 0.7317                                                             & 0.7805                                                              & 25               \\
Singapore         & 0.8710                                                              & 0.9032                                                               & 31               & Iraq                 & 0.7200                                                             & 0.8800                                                              & 24               \\
Angola            & 0.8696                                                              & 0.9130                                                               & 69               & Sudan                & 0.7083                                                             & 0.9167                                                              & 34               \\
Bangladesh        & 0.8679                                                              & 0.8679                                                               & 53               & Malaysia             & 0.7059                                                             & 0.7353                                                              & 34               \\
Botswana          & 0.8642                                                              & 0.8642                                                               & 81               & Spain                & 0.7059                                                             & 0.7941                                                              & 50               \\
Bahrain           & 0.8636                                                              & 0.9091                                                               & 66               & Honduras             & 0.7000                                                             & 0.8400                                                              & 43               \\
Cyprus            & 0.8636                                                              & 0.8788                                                               & 44               & Italy                & 0.6977                                                             & 0.7907                                                              & 43               \\
Austria           & 0.8630                                                              & 0.8904                                                               & 73               & Tunisia              & 0.6977                                                             & 0.8372                                                              & 33               \\
Ecuador           & 0.8615                                                              & 0.9231                                                               & 65               & Colombia             & 0.6970                                                             & 0.9394                                                              & 33               \\
Burundi           & 0.8594                                                              & 0.8750                                                               & 64               & France               & 0.6970                                                             & 0.8182                                                              & 33               \\
Estonia           & 0.8594                                                              & 0.8750                                                               & 64               & Mexico               & 0.6970                                                             & 0.8182                                                              & 56               \\
Unknown           & 0.8571                                                              & 0.8571                                                               & 100              & Belgium              & 0.6964                                                             & 0.7857                                                              & 36               \\
Serbia            & 0.8431                                                              & 0.8431                                                               & 51               & Chile                & 0.6944                                                             & 0.7778                                                              & 26               \\
Sweden            & 0.8409                                                              & 0.8864                                                               & 44               & Palestine            & 0.6923                                                             & 0.8077                                                              & 26               \\
India             & 0.8400                                                              & 0.8800                                                               & 50               & Syria & 0.6923                                                             & 0.8077                                                              & 25               \\
Kuwait            & 0.8333                                                              & 0.8667                                                               & 36               & Libya                & 0.6800                                                             & 0.8800                                                              & 53               \\
Lebanon           & 0.8333                                                              & 0.8333                                                               & 30               & Ghana                & 0.6792                                                             & 0.8302                                                              & 37               \\
Yemen             & 0.8333                                                              & 0.8750                                                               & 24               & Germany              & 0.6757                                                             & 0.7568                                                              & 45               \\
Maldives          & 0.8169                                                              & 0.8169                                                               & 71               & El Salvador          & 0.6667                                                             & 0.9111                                                              & 40               \\
Ireland           & 0.8163                                                              & 0.8980                                                               & 49               & Nigeria              & 0.6500                                                             & 0.7250                                                              & 57               \\
Norway            & 0.8163                                                              & 0.8571                                                               & 49               & Canada               & 0.6491                                                             & 0.8947                                                              & 25               \\
Iran              & 0.8158                                                              & 0.8684                                                               & 38               & Oman                 & 0.6400                                                             & 0.6800                                                              & 25               \\
Israel            & 0.8158                                                              & 0.8421                                                               & 38               & Saudi Arabia         & 0.6400                                                             & 0.7200                                                              & 28               \\
Panama            & 0.8140                                                              & 0.8605                                                               & 43               & Peru                 & 0.5625                                                             & 0.6875                                                              & 32               \\
Uruguay           & 0.8125                                                              & 0.8750                                                               & 32               &                      &                                                                    &                                                                     &                 
\end{tabular}
\caption{\label{country-accuracy}
Accuracy scores for the top performing word + character model across names from different countries.}
\end{table*}

\end{document}